%% file: main.tex
\documentclass[conference]{IEEEtran}
\IEEEoverridecommandlockouts
\usepackage{cite}
\usepackage{amsmath,amssymb,amsfonts}
\usepackage{algorithmic}
\usepackage{booktabs}

\usepackage{subfigure}
\usepackage{graphicx}
\usepackage{multirow}
\usepackage{textcomp}
\usepackage{xcolor}
\usepackage{soul}
\usepackage{overpic}
\usepackage{colortbl}

\RequirePackage{xspace}
\makeatletter
\DeclareRobustCommand\onedot{\futurelet\@let@token\@onedot}
\def\@onedot{\ifx\@let@token.\else.\null\fi\xspace}

\makeatother

\usepackage[pagebackref,breaklinks,colorlinks]{hyperref}

\usepackage{cleveref}

\usepackage{orcidlink}

\def\BibTeX{{\rm B\kern-.05em{\sc i\kern-.025em b}\kern-.08em
    T\kern-.1667em\lower.7ex\hbox{E}\kern-.125emX}}
\begin{document}

\title{A Temporal Modeling Framework for \\ Video Pre-Training on Video Instance Segmentation}

\author{
  \IEEEauthorblockN{
    \begin{tabular}{ccc}
      1\textsuperscript{st} Qing Zhong & 2\textsuperscript{nd} Peng-Tao Jiang\textsuperscript{*} & 3\textsuperscript{rd} Wen Wang \\
      University of Adelaide & vivo & Zhejiang University \\
      Adelaide, Australia & Hangzhou, China & Hangzhou, China \\[1ex] 
      4\textsuperscript{th} Guodong Ding & 5\textsuperscript{th} Lin Wu & 6\textsuperscript{th} Kaiqi Huang \\
      NUS & Swansea University & University of South China \\
      Singapore & Swansea, United Kingdom & Hengyang, China \\
    \end{tabular}
  }
}

\maketitle

\begin{abstract}
Contemporary Video Instance Segmentation (VIS) methods typically adhere to a pre-train then fine-tune regime, where a segmentation model trained on images is fine-tuned on videos. However, the lack of temporal knowledge in the pre-trained model introduces a domain gap which may adversely affect the VIS performance.  
To effectively bridge this gap, we present a novel ``\textit{video pre-training}'' approach to enhance VIS models, especially for videos with intricate instance relationships. Our crucial innovation focuses on reducing disparities between the pre-training and fine-tuning stages. 
Specifically, we first introduce consistent pseudo-video augmentations to create diverse pseudo-video samples for pre-training while maintaining the instance consistency across frames. Then, we incorporate a multi-scale temporal module to enhance the model's ability to model temporal relations through self- and cross-attention at short- and long-term temporal spans.
Our approach does not set constraints on model architecture and can integrate seamlessly with various VIS methods. Experiment results on commonly adopted VIS benchmarks show that our method consistently outperforms state-of-the-art methods. Our approach achieves a notable 4.0\% increase in average precision on the challenging OVIS dataset. 
\end{abstract}


\section{Introduction}
\label{sec:intro}

\begingroup
\renewcommand\thefootnote{\textsuperscript{*}}
\footnotetext{ indicates corresponding author (email: pt.jiang@vivo.com)}
\endgroup

Video Instance Segmentation (VIS) integrates classification, segmentation, and tracking of instances within a video sequence. VIS has been crucial for applications like autonomous driving~\cite{auto_1,auto_2}, video editing~\cite{video_edit_1,video_edit_2}, and video understanding~\cite{video_editing,ding2024coherent, ding2022temporal, ding2022leveraging, zhong2024onlinetas, survey}. VIS methods are generally categorized into two paradigms: online~\cite{masktrackrcnn,crossvis,minvis,idol,genvis,ctvis,dvis} and offline~\cite{vistr,mask2formervis,seqformer,vita,dvis}. Online methods segment video frames individually and track instances across frames, while offline approaches process the entire sequence simultaneously, integrating tracking and segmentation. Both paradigms typically follow a \emph{pre-train then fine-tune} strategy. They start with an image-trained segmentation model and then apply a temporal model for instance tracking.

The common \emph{pre-train then fine-tune} regime can lead to sub-optimal performance due to reliance on image-trained segmentation model applied directly to videos.  
A key challenge in video pre-training is the limited availability of suitable datasets, as popular datasets like YoutubeVIS~\cite{vis2021,vis} and OVIS~\cite{ovis} contain only a few thousand, or sometimes just a few hundred, videos, making effective pre-training difficult. Annotating larger video datasets is costly and time-consuming. 
Prior studies~\cite{ctvis,dvis,genvis,idol,vita} have generated pseudo-videos from static images using data augmentation, but they often fail to preserve temporal coherence.   
For example, CTVIS~\cite{ctvis} generates pseudo-video by combing same image augmented by random ``Copy \& Paste'' and ``Crop Shift'' independently. 
To address the above issue, we first introduce \textbf{consistent pseudo-video generation} that better simulates real instance motion by ensuring temporal consistency, particularly benefiting challenging video datasets with visually similar instances. 

When we generate the pseudo-video data using the proposed coherent pseudo-video generation strategy, we can train the image segmentation models to serve as 
a pertaining model for video instance segmentation.
However, image segmentation models directly pre-trained on the pseudo-video data 
may compromise effectiveness due to the lack of temporal awareness. 
These models are inherently static and incapable of handling the dynamic nature of video sequences, where objects enter and exit the field of view. To address this, various approaches~\cite{vistr,mask2formervis,seqformer,vita,dvis} have introduced temporal modules to capture instance relations over time. For instance, \cite{mask2formervis} extends the image segmentation model to video processing by introducing shared latent instance tokens. 
However, these models often struggle with scenarios involving intersecting temporal trajectories of multiple instances, highlighting a key limitation. 
In addition to the pseudo-video generation, we propose a \textbf{video pre-training strategy} that leverages a multi-scale temporal module (MSTM) to improve temporal processing. MSTM first captures local motion variations to enhance instance relations, then integrates these correlations across the sequence for a stronger temporal understanding. We empirically show that integrating MSTM during fine-tuning significantly boosts VIS model performance. 

\input{tex/fig_augimg}
\input{tex/fig_framework}

Our contributions are as follows:
\textbf{1)} We introduce the first video pre-training approach for VIS, bridging the gap between image pre-training and video fine-tuning. This method is easily integrated with various VIS models.
\textbf{2)} We develop a consistent pseudo-video generation technique that enriches training data diversity while maintaining consistency of video content.
\textbf{3)} We design a multi-scale temporal module to leverage both long- and short-term temporal correlations during pre-training and fine-tuning. 
\textbf{4)} Our framework achieves state-of-the-art performance on three standard VIS benchmarks and also improves image instance segmentation. 

\section{Related Work}
\label{sec:review}

\subsection{Online Video Instance Segmentation}

Online approaches~\cite{masktrackrcnn,crossvis,minvis,idol,genvis,ctvis,dvis} generate instance segmentations up to the latest frame without requiring future frames. These approaches typically add a temporal module for tracking, treating segmentation, and tracking separately. 
Amongst them, CTVIS~\cite{ctvis} builds on Mask2Former~\cite{mask2former} and incorporates the insight of online methods to learn more robust instance embeddings. 
In contrast, our work employs consistent pseudo-video augmentation during pre-training and integrates a multi-scale temporal module to better handle temporal data. 

\subsection{Data Augmentation}
Various augmentation techniques have been extensively explored in images~\cite{augseg,autoa_0,cutmix, cutmix_0,cutmix_1,gmflow} to enrich data sets and improve model training. 
Similar strategies have been applied in VIS during fine-tuning~\cite{augseg,randaug,autoa_0}, but they would often fall short in capturing the complex dynamics of real videos. 
CTVIS~\cite{ctvis} combines techniques such as copy-paste, cropping, and rotation, but using such augmentation lacks the capability to accurately simulate intricate instance interactions. 
To address this, we propose to enrich the pseudo-video diversity by combining various augmentation effects from a predefined pool. 
Our approach dynamically selects and reintegrates instances within the image, adding more realism compared to template-based methods. 

\section{Video Pre-training for VIS}

\subsection{Task Definition}
Formally, given an input video sequence $\mathbf{V}=\{I_t\}_{t=1}^{T}$, where $I_t$ represents the $t$-th frame in the video and $T$ is the total number of frames, the goal of video instance segmentation is to produce pixel-level segmentation masks for each object instance present in each frame. Let $\mathbf{O}=\{o_i\}_{i=1}^{N}$ denote the set of $N$ object instances across all frames, where $o_i$ represents an individual object instance. The objective is to segment each object instance $o_i$ into its constituent pixels, providing accurate spatial delineation throughout the video.

\subsection{Consistent Pseudo-Video Augmentation}
\label{subsec:augs}

A pseudo-video is temporally coherent when it maintains consistency in colour, semantic class, and spatial characteristics of instances across frames. 
We propose two augmentations: Consistent Stochastic (CoSt) and Video Morph \& Splice (VMoSp). CoSt uniformly adjusts the color and spatial attributes of individual frames within a pseudo-video sequence. VMoSp simulates complex video scenarios with objects of similar shapes, sizes, and colors, often involving occlusions.

\textbf{Consistent Stochastic Augmentation.} CoSt randomly selects and blends augmentation strategies used in~\cite{augseg} to alter the colour and spatial features of video frames. 
Specifically, unlike previous works applying different blending of augmentations to different frames, we utilize the same blending of the selected augmentations across all the frames, which ensures smooth transitions between frames. 
%
We randomly extract up to $k=3$ augmentations from the pool. 

\textbf{Video Morph \& Splice Augmentation.} In videos, instances moving in opposite directions can cause occlusions or sudden position changes, leading to tracking errors like ID switching. We address this with the VMoSp.  
First, we replicate an image $I$ for $T$ times, applying simple augmentation like random rotations to create a naive pseudo-video $V$. Next, we select an instance $o$ and its mask $m$ from $I$, and isolate $o$ to form $o^I = I \odot m$. Then, we apply minor morph augmentations (flips, scaling, and rotations) to $o^I$ and $m$ for each frame, denoted as $A_t$ for the $t^{th}$ frame. Finally, we splice the $T$ augmented frames into the pseudo-video $V$. Frame $V_t$ in the spliced video is written as:
\begin{align}
V_t = V_t\times (1 - A_t(m))+A_t(o^I)
\end{align}
If the spliced instance occludes an instance with original mask $\hat{m}$ in $t^{th}$ frame, the mask $\hat{m}_t$ of the occluded instance becomes $\hat{m}_t = \hat{m} \times (\mathbf{1} - A_t(m))$. 
This procedure improves the consistency of instances, assisting in developing robust VIS models.

\textbf{Disucssion.} 
A straightforward approach is employing Copy \& Paste or CutMix-based techniques, adapted from image segmentation, for augmenting videos~\cite{copypaste, cutmix, cutmix_0, cutmix_1}. 
Nonetheless, applying image-based augmentation separately to each video frame challenges consistency across consecutive frames. As depicted in the top and second rows of~\cref{fig:aug_img}, using image copy \& paste to augment pseudo-videos by inserting different instances into frames can lead to semantic anomalies, such as an airplane suddenly appearing in one frame and then disappearing abruptly in the next, substituted by a monkey. Such inconsistencies are less common in real-world videos, and extensive training on these inconsistent pseudo-videos may lead to degraded performance. With our proposed approach, depicted in the bottom row, minimal changes occur between adjacent frames, ensuring instances do not abruptly disappear or appear. Furthermore, it simulates complex video scenarios featuring multiple instances of similar colour, size, and semantic class (the augmented dog in green). 

\subsection{Multi-Scale Temporal Module}
\label{sec: mstm}

We next present our multi-scale temporal module to address the disparity of using a image-pretrained model to initialise video temporal segmentation model. MSTM enhances temporal information by reinforcing correlations between adjacent frames and aggregating  long-term relations. 
It operates on multi-scale frame features extracted by Transformer layers in the temporal encoder. For each feature scale $s$, we split $f_s \in \mathbb{R}^{T \times H \times W \times D}$ along the temporal dimension into $f^1_s$ and $f^2_s \in \mathbb{R}^{\frac{T}{2} \times H \times W \times D}$, where $T, H, W,$ and $D$ are the number of frames, height, width, and feature dimension, respectively. 
The split features are recombined as $f_s^{'} = \text{concat}(f^2_s, f^1_s)$. 
We then apply the following SWMA (shift window based multi-head attention) operation:
\begin{align}
z_s^l = \text{SWMA}(f_s, f_s^{'}) + z_s^{l-1},
\label{eq:swin_block}
\end{align}
where $z^l$ represents the output features and $l$ is the layer index. The self-attention block processes $f_s$ and $f_s^{'}$ independently, while the frame-wise cross-attention block uses $f_s$ as the query and $f_s^{'}$ as the key and value. The process is repeated $L$ times for short-term temporal interactions.  
The output $z^L$ from \cref{eq:swin_block} is then passed through $N$ ConvGRU~\cite{rvm} layers for long-term temporal interactions across the entire video sequence. The original feature $f_s$ is combined with the ConvGRU output to generate the final output. The multi-scale features from MSTM are then sent to an FPN block, which feeds into the Mask2Former decoders for instance segmentation predictions. The overall structure is illustrated in~\cref{fig:method}. Our learning objective aligns with that of Mask2Former~\cite{mask2former}.

\input{tex/abl_rot}

\section{Experiments}
\subsection{Datasets \& Evaluation Metrics}
\label{subsec:datatsets}
We evaluate our video pre-training approach on three VIS benchmarks:
\textbf{YouTubeVIS-2019}\cite{vis} with 2,238 training videos, 302 validation videos, and 343 test videos across 40 categories;
\textbf{YouTubeVIS-2021}\cite{vis2021}, an extension with 2,985 training and 453 validation videos;
\textbf{OVIS}~\cite{ovis} with 607 training, 140 validation, and 154 test videos. 
Following~\cite{ctvis,genvis,dvis,vita}, we report Average Precision (AP) and Average Recall (AR).

\subsection{Implementation Details}
\textbf{Pseudo-video Generation.} 
We generate pseudo-videos as described in \cref{subsec:augs} with COCO~\cite{mscoco} images for pre-training. We then use pseudo-videos from both the COCO and VIS datasets to fine-tune the VIS model, with sequences of three and ten frames, respectively, as per the VIS baseline~\cite{ctvis}.
 
\input{tex/fig_tsne}

\textbf{VIS Model.} We opt for the state-of-the-art approach CTVIS~\cite{ctvis} as our baseline VIS model. CTVIS builds upon pre-trained Mask2Former~\cite{mask2former} instance segmentation approach based on Resnet-50~\cite{resnet} and Swin-L~\cite{swin}, and enhancing feature compactness via contrastive learning. Despite these advancements, CTVIS maintains the use of image pre-training. 
Our pseudo-video augmentation module operates directly on input data without imposing constraints on instance segmentation models. For the pre-training instance segmentation model, Mask2Former~\cite{mask2former} is used. We preserve their original settings without any modifications to ensure a fair comparison with other state-of-the-art methods.  

\subsection{Ablation Studies}

\input{tex/tab_augs}

\input{tex/tab_main}

\textbf{Module Effectiveness.} Using CTVIS~\cite{ctvis} as our baseline, we integrate and evaluate our proposed modules and report results in~\cref{tab:allabl}.  
Adding MSTM improves AP by 0.1\% and 0.9\% on two benchmarks, demonstrating its value for capturing temporal relations. 
Incorporating CPVA further enhances performance, indicating the importance of consistent video augmentations. 
Combining video pre-training (VP) shows a 3.7\% AP improvements on OVIS and 0.7\% on YIS19, reflecting the added complexity of OVIS and the benefits of our VP. The best results are achieved when all modules are combined, improving AP by 2.1\% and 5.2\% over baselines. 

\input{tex/fig_qual_result}
\textbf{Augmentation Types.} 
\cref{tab:rotation} shows that a small rotation angle ($\pm15^\circ$) yields the best performance, with 59.7\% AP on YoutubeVIS-2019 and 27.4\% AP on OVIS. Larger rotation angles decrease performance by $\sim$1\%, likely due to significant positional change between frames. 

\cref{tab:augmentation} compares our approach with existing methods. Crop Shift~\cite{ctvis} achieves the lowest performance across both datasets, highlight its failure to ensure temporal consistency. Autoaug, which applies random transformations, shows similar performance to Copy \& Paste (59.5\% vs. 59.1\% in AP).
Our proposed augmentations achieves the best results: 
VMoSp reaches at 27.6\% on the OVIS, while CoSt achieves 60.0\% on the YoutubeVIS-19. This indicates that maintaining temporal consistency is crucial for less complex videos (YoutubeVIS-19), whereas enhancing diversity is key for more complex videos (OVIS). 

\input{tex/tab_insabl}

\textbf{VMoSp Instances.} VMoSp employs uniform sampling to select one instance in the \emph{current} frame to augment the pseudo-video. We first compare it to Copy \& Paste, which augments the pseudo-video with instances sampled from \emph{arbitrary} frames. As indicated in~\cref{tab:insabl}, when employing a single instance for augmentation, VMoSp exhibits notable improvements of 1.4\% in AP over Copy \& Paste on the more challenging dataset OVIS. This demonstrates the effectiveness of our approach in simulating more intricate scenarios in videos. 

We further increase the number of instances selected for augmentation. Results are also presented in~\cref{tab:insabl}. An overall improvement is observed when employing two instances to augment the pseudo-video. However, with a maximum of 3 instances, there is a slight performance drop, yet it remains comparable and competitive compared to the Copy \& Paste method (31.9\% vs. 31.2\% on OVIS). 

\textbf{Query Embedding Visualisation.} We utilise T-SNE \cite{tsne} to visualise the instance feature space after image and video pre-training, respectively, to observe the impact of our approach on feature learning. 
In~\cref{fig:tsne}, we plot the instance features from two distinct videos for each benchmark. It is evident that features representing the same instance across different frames, denoted by the same colour, are scattered in the feature space with image pre-training (the top row). Conversely, our video pre-training efficiently clusters identical instance features across frames (the bottom row). This enhanced compactness is primarily attributed to our specially designed MTSM module, which effectively harnesses the temporal information for superior feature representation learning. Therefore, we emphasise incorporating temporal aspects during pre-training for downstream video-related tasks.  

\subsection{Comparison to the State-of-the-art}
\cref{tab:main} compares our video pre-training approach with SOTA methods that rely on image pre-training. Our method, using ResNet-50, consistently outperforms leading VIS methods like CTVIS~\cite{ctvis} and GenVIS~\cite{genvis}, demonstrating the limitations of image pre-training in fully leveraging temporal modules. 
The two-stage method DVIS~\cite{dvis}, also with ResNet-50, underperforms by 3.4\% to 5.7\% AP across benchmarks due to its inadequate tracking in the first stage. Our approach, by training a robust video pre-trained model, achieves superior results and shows significant improvements of 0.9\% to 4.0\% AP on more complex benchmarks compared to CTVIS~\cite{ctvis}, and performs competitively with Swim-L~\cite{swin}.


\subsection{Qualitative Results}
We next provide a qualitative analysis of our approach, presented in~\cref{fig:qual_result}, compared against the performance of CTVIS~\cite{ctvis} and IDOL~\cite{idol}. Our model accurately tracks all object instances across the video frames, as evidenced by the bottom row. In contrast, both IDOL~\cite{idol} and CTVIS~\cite{ctvis} suffer from common issues such as partial overlapping, over- and under-segmentation, as seen in the left example, and missed instances in the right example. 
The shortcomings of these models stem primarily from the interference between the segmentation and temporal modules. Additionally, they encounter difficulties detecting instances due to severe occlusions or instances disappearing and reappearing.

\input{tex/fig_fial}

\textbf{Failure Cases.} We illustrate some cases where our approach in~\cref{fig:failure_cases} faces challenges, including classification errors, ID switches, and undetected instances in YouTubeVIS-2019 videos. These issues predominantly stem from the long-tail problem in the dataset. Moreover, the current annotations in VIS datasets are sparse, typically every five frames, resulting in significant variations between adjacent frames, making the task more challenging.  

\input{tex/tab_instseg}
\subsection{Instance Segmentation}
We additionally evaluate our approach on the instance segmentation task using the COCO dataset~\cite{mscoco} and present the performance results in~\cref{tab:instseg}. We replace the original image pre-training with our proposed video pre-training for Mask2Former~\cite{mask2former} and Mask DINO~\cite{maskdino} using ResNet-50~\cite{resnet} and Swin-L~\cite{swin} as backbones when applicable. It is evident that our proposed approach consistently outperforms the baseline methods across all metrics. This reinforces the effectiveness of our pseudo-video augmentation technique in enhancing sample diversity.

\section{Conclusion}
This paper presents video pre-training for video instance segmentation tasks to rectify the often neglected misalignment between image pre-training and video fine-tuning. We introduce consistent pseudo-video augmentations to create pseudo-video to simulate real-world variations alongside a multi-scale temporal module implemented to better exploit temporal information across frames. These innovations serve to close the disparity between pre-training and fine-tuning stages. 
As a result, significant performance improvements have been observed across VIS benchmarks. Furthermore, we demonstrate that video pre-training can effectively enhance the performance of instance segmentation tasks on images. 
In the future, we aim to expand pseudo-video augmentations to cover a broader range of real-world scenarios, enhancing the robustness of our model. 

\bibliographystyle{IEEEtran}
\bibliography{main}

\end{document}


\title{A Temporal Modeling Framework for \\ Video Pre-Training on Video Instance Segmentation Supplementary}

\author{Anonymous ICME submission}

\maketitle


\section{Cost augmentation pool}
The CoSt augmentation builds upon the augmentation pool proposed by~\cite{augseg} that randomly selects and combines various strategies up to $k=3$ times to modify the colour and spatial characteristics of each frame in the pseudo-video.  \cref{tab:pool} lists all augmentation strategies. 
\input{tex/tab_pool}

\section{Video Pre-trained VIS Model}
In our experiment, we mainly adopted Mask2Former~\cite{mask2former} as the segmentation model due to its superior performance in VIS. It is interesting to mention that the superior performance of instance segmentation methods doesn't necessarily translate to better performance on VIS tasks. As we show in~\cref{tab:ins_baseline}, for instance segmentation task on COCO~\cite{mscoco}, Mask DINO~\cite{maskdino} surpasses Mask2Former by 1.9\% in AP. However, when adopted to VIS, Mask2Former, serving as the segmentation model, outperforms Mask DINO by 1.3\% in AP on the YouTubeVIS-2019~\cite{vis} benchmark. 

\input{tex/tab_insbaseline}

\section{MSTM in pre-training}
We conduct additional experiments by removing the short- and long-term sub-modules in our proposed MSTM and present the results in ~\cref{tab:mstt_pretrain}. As inputs, we simply duplicate the image three times without any data augmentation. As we can see, incorporating long-term or short-term temporal sub-modules each enhances performance by 0.9\% and 0.8\% in AP. While combing both sub-modules in MSTM achieves the overall best performance, highlighting the importance of taking into consideration temporal interactions across varied spans. 
\input{tex/tab_coco}

\section{Visualisation Results} \label{app:visual}

\begin{figure*}[t]
    \centering
\includegraphics[width=2.0\columnwidth]{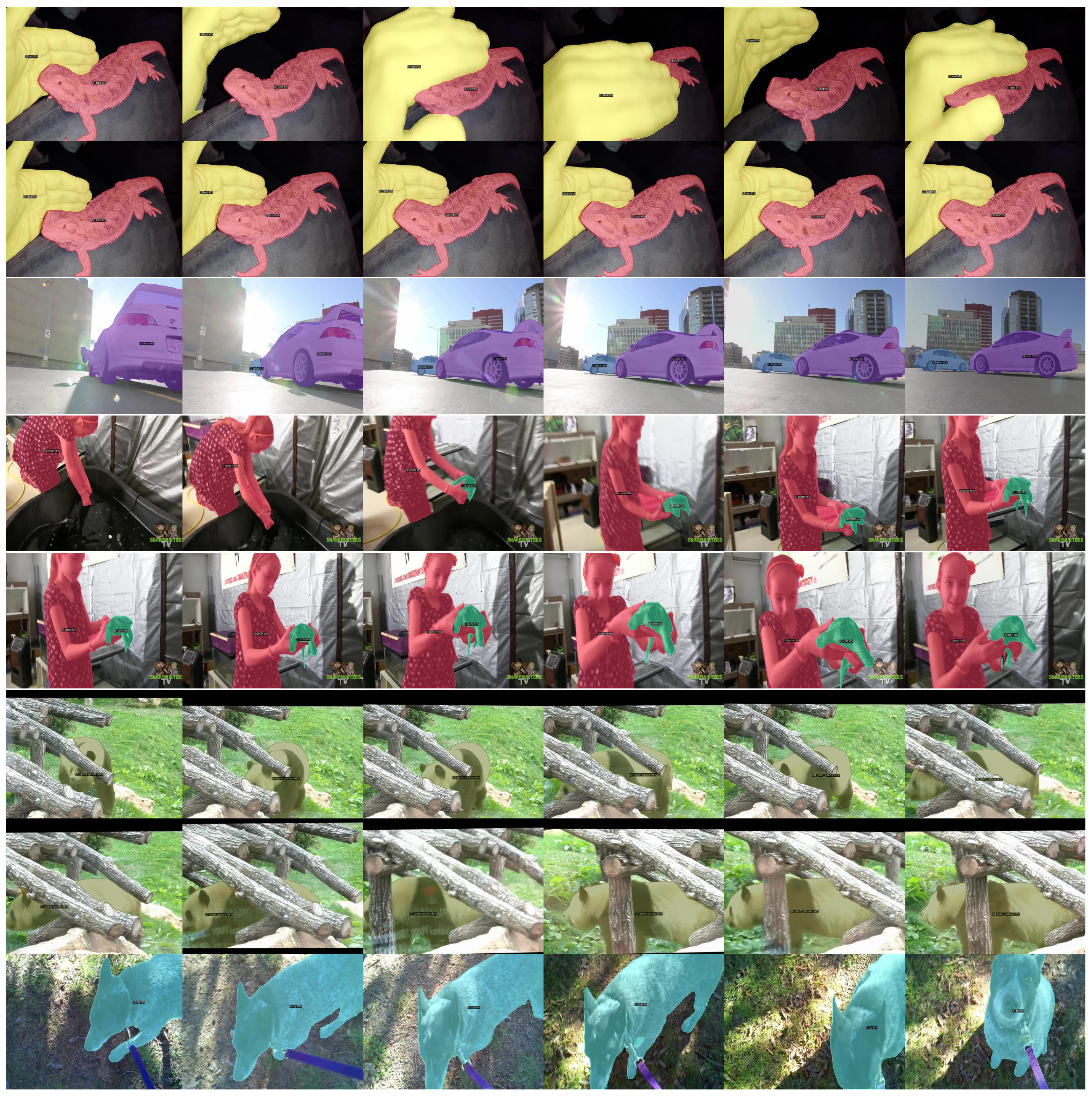}
\caption{YouTubeVIS-2019~\cite{vis} Visualisation Results}
    \label{fig:19_vis}
\end{figure*}

\begin{figure*}[t]
    \centering
\includegraphics[width=2.0\columnwidth]{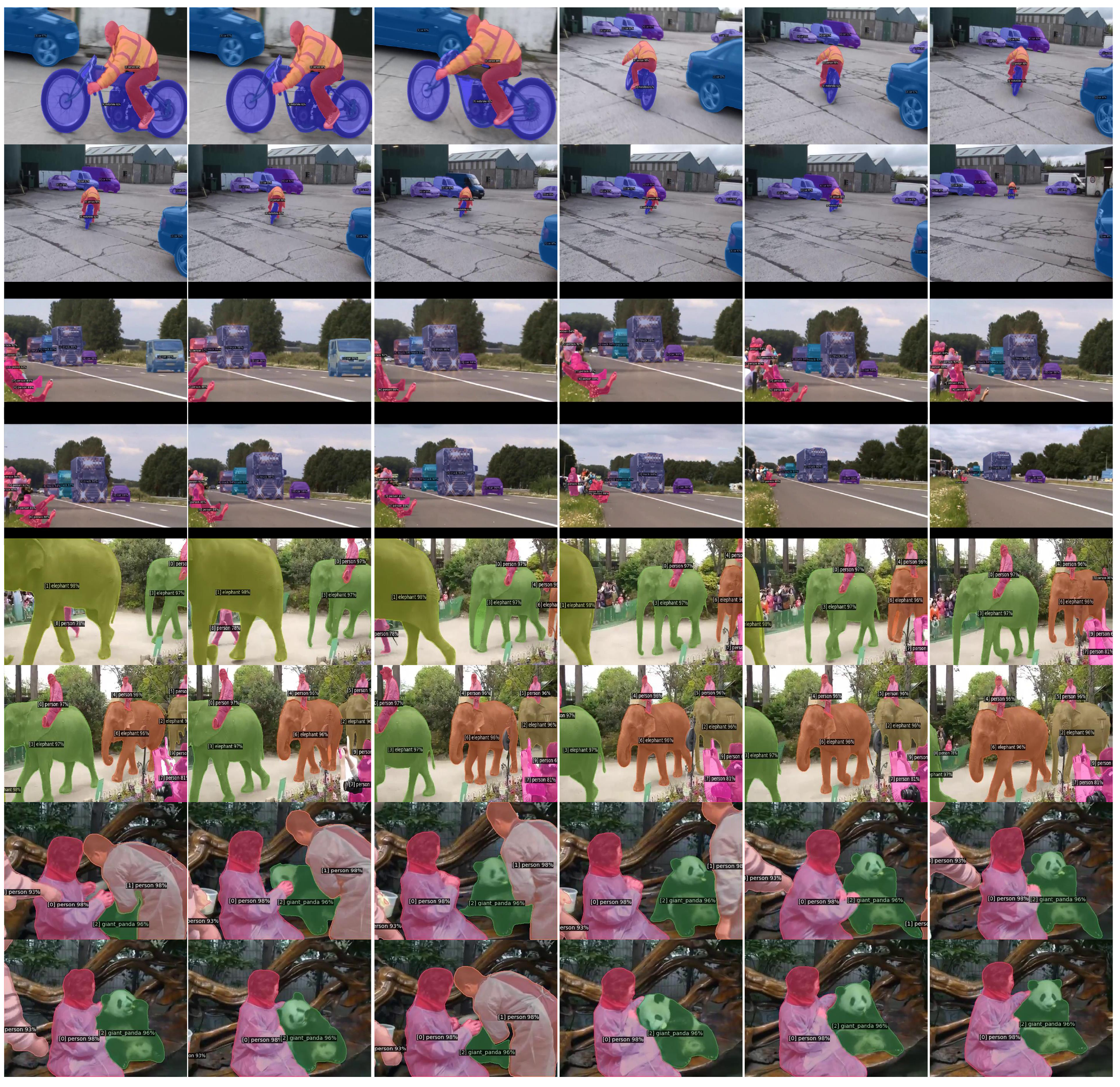}
\caption{YouTubeVIS-2021~\cite{vis2021} Visualisation Results}
    \label{fig:21_vis}
\end{figure*}

\begin{figure*}[t]
    \centering
\includegraphics[width=2.0\columnwidth]{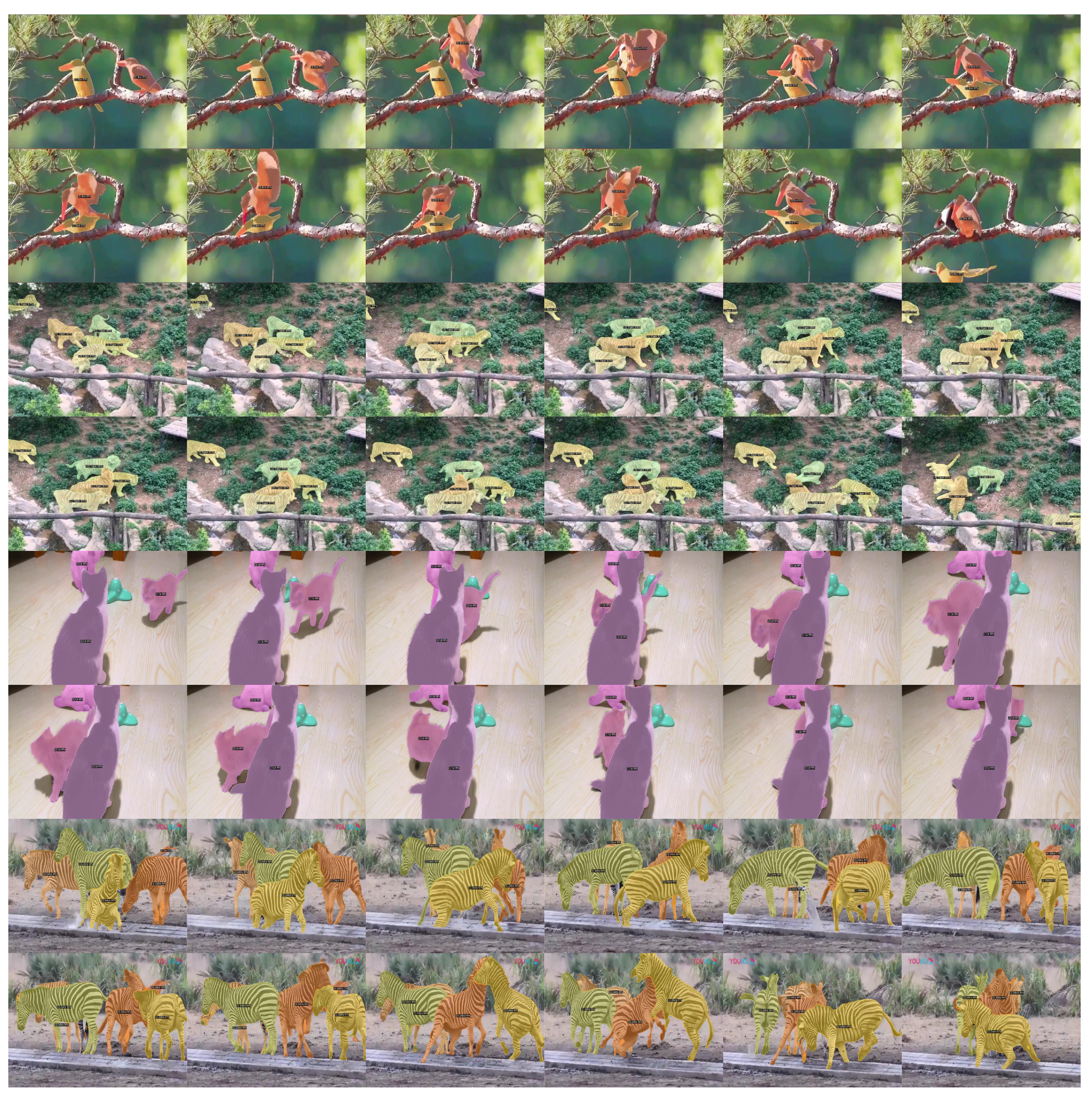}
\caption{OVIS~\cite{ovis} Visualisation Results}
    \label{fig:ovis}
\end{figure*}

\cref{fig:19_vis}, \cref{fig:21_vis}, and \cref{fig:ovis} present qualitative results on the three major benchmarks YoutubeVIS-2019~\cite{vis}, YoutubeVIS-2021~\cite{vis2021} and OVIS~\cite{ovis}, respectively. Despite the presence of highly intricate scenes and motions in these challenging video sequences, our approach exhibits great capability in accurately segmenting and tracking objects of interest. The visual examples clearly demonstrate the efficacy of our method in handling complex real-world scenarios.

\bibliographystyle{IEEEbib}
\bibliography{icme2025references}

%% file: tex/fig_augimg.tex
\begin{figure}[t]
    \centering
    \includegraphics[width=0.8\linewidth]{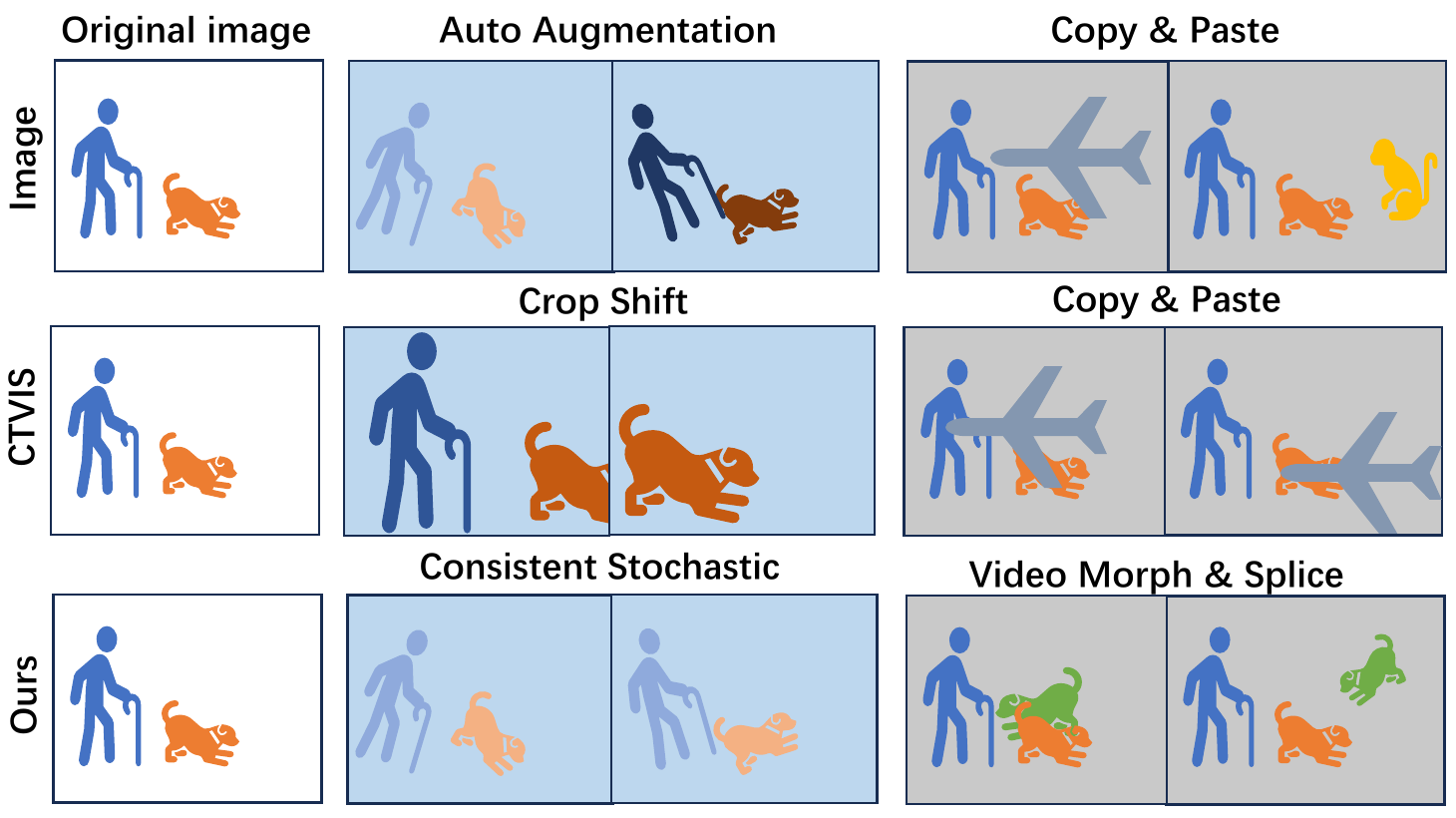}
    \caption{Augmentation comparison. 
    }
    \label{fig:aug_img}
    \vspace{-5mm}
 \end{figure}

%% file: tex/fig_framework.tex
\begin{figure*}[!t]

\centering
\begin{overpic}[width=1.8\columnwidth]{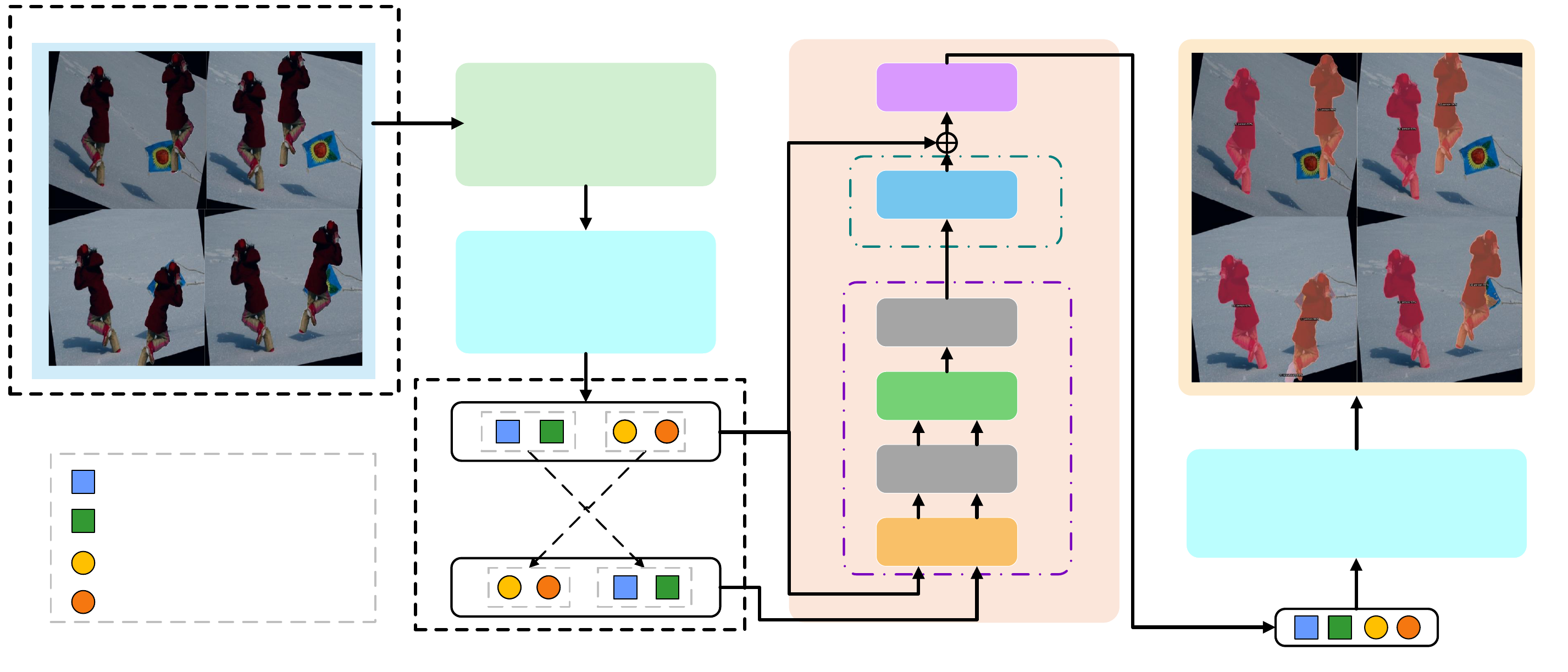}
\put(10,11.5){\scriptsize $1^{st}$ frame feature}
\put(10,9){\scriptsize $2^{nd}$ frame feature}
\put(10,6.5){\scriptsize $3^{rd}$ frame feature}
\put(10,4){\scriptsize $4^{th}$ frame feature}
\put(8,41){\scriptsize Pseudo-Video ($V$)}

\put(34.5,34.3){\scriptsize Backbone}
\put(34,25){\scriptsize Transformer}
\put(35,22.5){\scriptsize Encoder}

\put(29,11){\scriptsize Temporal}
\put(30,9.5){\scriptsize Swap}

\put(29,14.7){\tiny $f_s^1$}
\put(43.7,14.7){\tiny $f_s^2$}
\put(58.6,4.5){\tiny $f_s$}

\put(56.4,14.7){\tiny $Q$}
\put(62.7,14.7){\tiny $K,V$}

\put(58,25.5){\tiny $z_s^l$}

\put(29,5){\tiny $f_s^2$}
\put(44.1,5){\tiny $f_s^1$}
\put(62.6,4.5){\tiny $f_s^{'}$}

\put(59,41){\scriptsize MSTM}
\put(85,41){\scriptsize Output}
\put(59,36.8){\textcolor{white}{\scriptsize FPN}}
\put(57.3,29.8){\textcolor{white}{\scriptsize ConvGRU}}
\put(57,22){\scalebox{0.8}{\scriptsize \textcolor{white}{Norm \& Add}}}
\put(57.7,17.3){\scalebox{0.8}{\scriptsize \textcolor{white}{Swin Cross}}}
\put(57,12.5){\scalebox{0.8}{\scriptsize \textcolor{white}{Norm \& Add}}}
\put(57.7,8){\scalebox{0.8}{\scriptsize \textcolor{white}{Swin Self}}}

\put(65.5,7){\scriptsize $L \times$}
\put(63.8,27.5){\scriptsize $N \times$}

\put(80.3,10.5){\scriptsize Transformer Decoder}
\end{overpic}
\vspace{-2mm}
\caption{Overview of VIS model integrated with multi-scale temporal module. 
}
\label{fig:method}
\vspace{-5mm}
\end{figure*} 

%% file: tex/abl_rot.tex
\begin{table}[t]
  \centering
  \begin{minipage}{0.51\linewidth}
    \caption{Ablation study on modules.}
    \label{tab:allabl}
    \centering
    \resizebox{\textwidth}{!}{%
    \begin{tabular}{ccccc}
    \toprule
    MSTM & CPVA & VP & AP$^{\mathtt{YV19}}$ & AP$^{\mathtt{OVIS}}$ \\
    \midrule
    - & - & - & 59.7 & 27.4 \\
    $\checkmark$ & - & - & 59.8 & 28.3 \\
    $\checkmark$ & $\checkmark$ & - & 60.2 & 28.5 \\
    $\checkmark$ & - & $\checkmark$ & 60.5 & 32.0 \\
    \rowcolor{gray!30}  
    $\checkmark$ & $\checkmark$ & $\checkmark$ & \textbf{61.8} & \textbf{32.6} \\
    \bottomrule
    \end{tabular}%
    }
  \end{minipage}
  \hfill
  \begin{minipage}{0.44\linewidth}
    \caption{Rotation angle settings.}
    \label{tab:rotation}
    \centering
    \begin{tabular}{lcc}
    \toprule
    Rotation & AP\textsuperscript{YV19} & AP\textsuperscript{OVIS} \\
    \midrule
    \rowcolor{gray!30}
    $\pm15^\circ$ & 59.7 & 27.4 \\
    $\pm45^\circ$ & 57.8 & 26.4 \\
    $\pm60^\circ$ & 58.8 & 27.2 \\
    \bottomrule
    \end{tabular}
  \end{minipage}
  \vspace{-5mm}
\end{table}

%% file: tex/fig_tsne.tex
\begin{figure*}[tb]
    \centering
    \includegraphics[width=1.8\columnwidth]{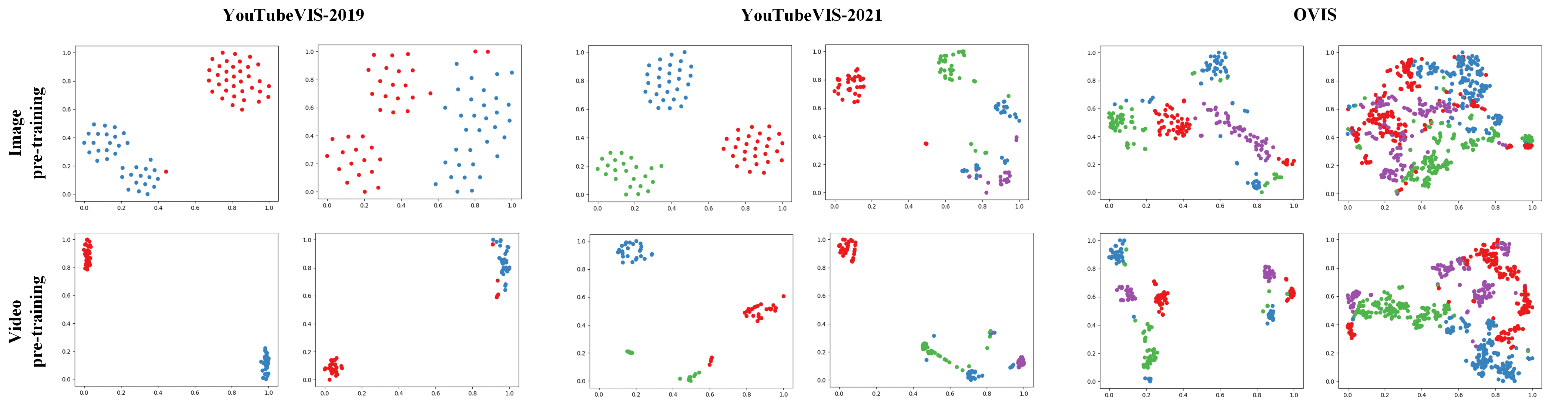}
    \caption{Visualization of query embeddings posts pre-training with images and videos using T-SNE.}
    \label{fig:tsne}
    \vspace{-10pt}
\end{figure*}

%% file: tex/tab_augs.tex
\begin{table}[t]
  \centering
  \caption{Different augmentation methodS.}
  \begin{tabular}{lcc}
    \toprule
    Augmentations & AP\textsuperscript{YV19} & AP\textsuperscript{OVIS} \\
    \midrule
    N/A &  54.1 & 24.5 \\
    Rotation & 59.7 & 27.4 \\
    Copy \& Paste & 59.1 & 26.8 \\
    \rowcolor{gray!30}
    Crop Shift & 58.6 & 26.1 \\
    Autoaug & 59.5 & 26.8 \\
    \rowcolor{gray!30}
    CoSt & 60.0 & 27.4 \\
    VMoSp & 59.8 & 27.6 \\
    \bottomrule
  \end{tabular}
  \label{tab:augmentation}
  \vspace{-5mm}
\end{table}

%% file: tex/tab_main.tex
\begin{table*}[t]
  \centering
      \caption{Performance comparison to SOTA methods. The best and second best are highlighted in \textbf{bold} and \underline{underlined}, respectively. 
    }
    \label{tab:main}%
    \resizebox{\textwidth}{!}{%
    \begin{tabular}{lcccccccccccccccc}
    \toprule
    \multirow{2}[1]{*}{Method} &\multirow{2}[1]{*}{Arch.} & \multicolumn{5}{c}{YouTubeVIS-2019\cite{vis}}                               & \multicolumn{5}{c}{YouTubeVIS-2021\cite{vis2021}}                               & \multicolumn{5}{c}{OVIS\cite{ovis}} \\ \cmidrule(lr){3-7}\cmidrule(lr){8-12}\cmidrule(lr){13-17}
    \multicolumn{2}{c}{}   & AP          & AP$_{\mathtt{50}}$        & AP$_{\mathtt{75}}$        & AR$_{\mathtt{1}}$         & AR$_{\mathtt{10}}$        & AP          & AP$_{\mathtt{50}}$        & AP$_{\mathtt{75}}$        & AR$_{\mathtt{1}}$         & AR$_{\mathtt{10}}$        & AP          & AP$_{\mathtt{50}}$        & AP$_{\mathtt{75}}$        & AR$_{\mathtt{1}}$         & AR$_{\mathtt{10}}$ \\ 
    \midrule
     MaskTrack \cite{masktrackrcnn}&\multirow{10}{*}{\rotatebox{90}{ResNet-50~\cite{resnet}}}  & 30.3        & 51.1        & 32.6        & 31.0          & 35.5        & 28.6        & 48.9        & 29.6        & 26.5        & 33.8        & 10.8        & 25.3        & 8.5         & 7.9         & 14.9 \\
     Mask2Former~\cite{mask2formervis}   &       & 46.4        & 68.0          & 50.0          & -           & -           & 40.6        & 60.9        & 41.8        & -           & -           & 17.3        & 37.3        & 15.1        & 10.5        & 23.5 \\
    SeqFormer \cite{seqformer}& & 47.4        & 69.8        & 51.8        & 45.5        & 54.8        & 40.5        & 62.4        & 43.7        & 36.1        & 48.1        & 15.1        & 31.9        & 13.8        & 10.4        & 27.1 \\
    MinVIS \cite{minvis}       &  & 47.4        & 69.0          & 52.1        & 45.7        & 55.7        & 44.2        & 66.0          & 48.1        & 39.2        & 51.7        & 25.0          & 45.5        & 24.0          & 13.9        & 29.7 \\
    IDOL \cite{idol}   && 49.5        & 74.0          & 52.9        & 47.7        & 58.7        & 43.9        & 68.0          & 49.6        & 38.0          & 50.9        & 30.2        & 51.3        & 30.0          & 15.0          & 37.5 \\
    VITA \cite{vita}& & 49.8        & 72.6        & 54.5        & 49.4        & 61.0          & 45.7        & 67.4        & 49.5        & 40.9        & 53.6        & 19.6        & 41.2        & 17.4        & 11.7        & 26.0 \\
    GenVIS \cite{genvis} & & 51.3 & 72.0 & 57.8 & 49.5 & 60.0 & 46.3 & 67.0 & 50.2 & 40.6 & 53.2 & \underline{35.8} & \underline{60.8} & \underline{36.2} & \underline{16.3} & 39.6 \\

    DVIS \cite{dvis} && 52.6 & 76.5 & 58.2 & 47.4 & 60.4 & 47.4 & 71.0 & 51.6 & 39.9 & 55.2 & 33.8 & 60.4 & 33.5 & 15.3 & 39.5 \\
                
    CTVIS  \cite{ctvis} &     & \underline{55.1} & \underline{78.2} & \underline{59.1} & \textbf{51.9} & \underline{63.2} & \underline{50.1} & \underline{73.7} & \underline{54.7} & \underline{41.8} & \underline{59.5} & 35.5 & \underline{60.8} & 34.9 & 16.1 & \underline{41.9} \\
    \rowcolor{gray!30}  
    \textbf{Ours} & &\textbf{56.0} & \textbf{78.6} & \textbf{60.8} & \underline{51.3} & \textbf{63.4} & \textbf{53.4} & \textbf{75.9} & \textbf{57.6} & \textbf{45.2} & \textbf{61.4} & \textbf{39.5} & \textbf{65.4} & \textbf{39.1} & \textbf{17.4} & \textbf{45.2} \\
    \midrule
    
    SeqFormer \cite{seqformer} & \multirow{9}{*}{\rotatebox{90}{Swin-L~\cite{swin}}} & 59.3        & 82.1        & 66.4        & 51.7        & 64.6        & 51.8        & 74.6        & 58.2        & 42.8        & 58.1        & -               & -           & -           & -           & - \\
    Mask2Former \cite{mask2formervis} &   & 60.4        & 84.4        & 67.0          & -           & -           & 52.6        & 76.4        & 57.2        & -           & -           & 25.8        & 46.5        & 24.4        & 13.7        & 32.2 \\
    MinVIS \cite{minvis}   & & 61.6        & 83.3        & 68.6        & 54.8        & 66.6        & 55.3        & 76.6        & 62          & 45.9        & 60.8        & 39.4        & 61.5        & 41.3        & 18.1        & 43.3 \\
    VITA \cite{vita} & & 63.0          & 86.9        & 67.9        & 56.3        & 68.1        & 57.5        & 80.6        & 61.0          & 47.7        & 62.6        & 27.7        & 51.9        & 24.9        & 14.9        & 33.0 \\
    IDOL \cite{idol}  &     & 64.3  & 87.5   & 71.0 & 55.5 & 69.1        & 56.1        & 80.8   & 63.5 & 45.0          & 60.1        & 42.6       & 65.7   & 45.2    & 17.9        & 49.6 \\

    GenVIS \cite{genvis} & &64.0 & 84.9 & 68.3 & 56.1 & 69.4 & 59.6 & 80.9 & 65.8 & 48.7 & 65.0 & 45.4 & 69.2 & 47.8 & 18.9 & 49.0 \\

    DVIS \cite{dvis} & &\underline{64.9} & \underline{88.0} & \textbf{72.7} & \underline{56.5} & 70.3 & 60.1 & 83.0 & 68.4 & 47.7 & 65.7 & \underline{48.6} & \textbf{74.7} & \underline{50.5} & 18.8 & \underline{53.8} \\ 
                
    CTVIS \cite{ctvis} & & \textbf{65.6} & 87.7 & \underline{72.2} & \underline{56.5} & \underline{70.4} & \underline{61.2} & \textbf{84.0} & \underline{68.8} & \underline{48.0} & \underline{65.8} & 46.9 & 71.5 & 47.5 & \underline{19.1} & 52.1 \\
    \rowcolor{gray!30}         
    \textbf{Ours} & & \textbf{65.6} & \textbf{88.3} & 71.8 & \textbf{57.0} & \textbf{71.3} & \textbf{62.2} & \underline{83.1} & \textbf{69.1} & \textbf{48.8} & \textbf{67.1} & \textbf{49.4} & 72.9 & \textbf{52.5} & \textbf{20.1} & \textbf{54.2} \\
    \bottomrule
    \end{tabular}%
    }
    \vspace{-3mm}
\end{table*}%

%% file: tex/fig_qual_result.tex
\begin{figure}[t]
    \centering
    \includegraphics[width=1.0\linewidth]{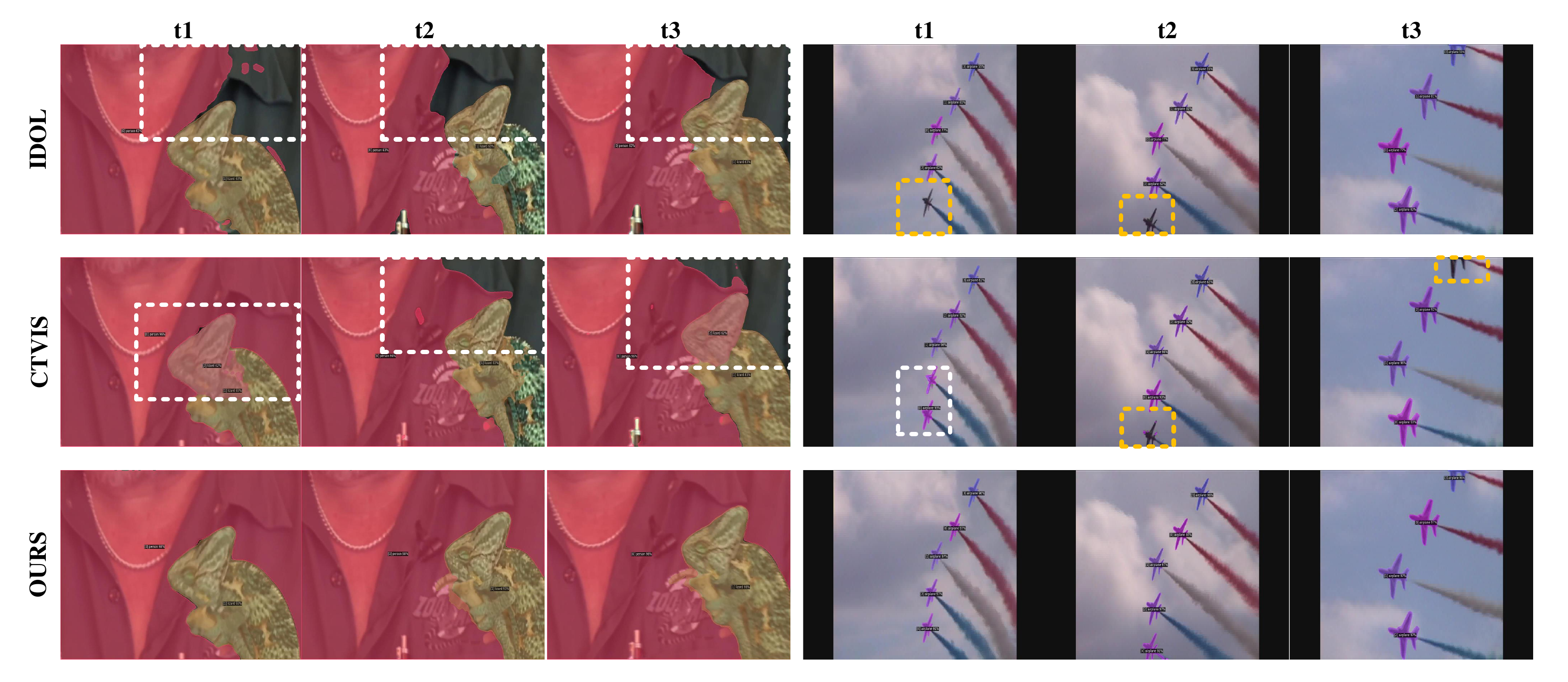}
    \caption{Qualitative results on YouTubeVIS-2019 \cite{vis}.}
    \label{fig:qual_result}
    \vspace{-3mm}
 \end{figure}

%% file: tex/tab_insabl.tex
\begin{table}[t]
  \caption{Ablation of the instance source and variations in the number of sampled instances for augmentation.}
  \label{tab:insabl}
  \centering
    \begin{tabular}{llccc}
    \toprule
    Method &Source&No. instances &AP$^{\mathtt{YV19}}$ &AP$^{\mathtt{OVIS}}$    \\
    \midrule
    Copy \& Paste& Arbitrary&      1     & 62.0 & 31.2  \\
    \rowcolor{gray!30}
    VMoSp & Current & 1  & 61.8 & 32.6\\
    VMoSp&Current & 2& 62.0 & 32.9 \\
    \rowcolor{gray!30}
    VMoSp&Current & 3& 61.9 & 31.9 \\
    \bottomrule
    \end{tabular}%
    \vspace{-5mm}
\end{table}

%% file: tex/fig_fial.tex
\begin{figure}[ht]
    \centering
    \includegraphics[width=1.0\linewidth]{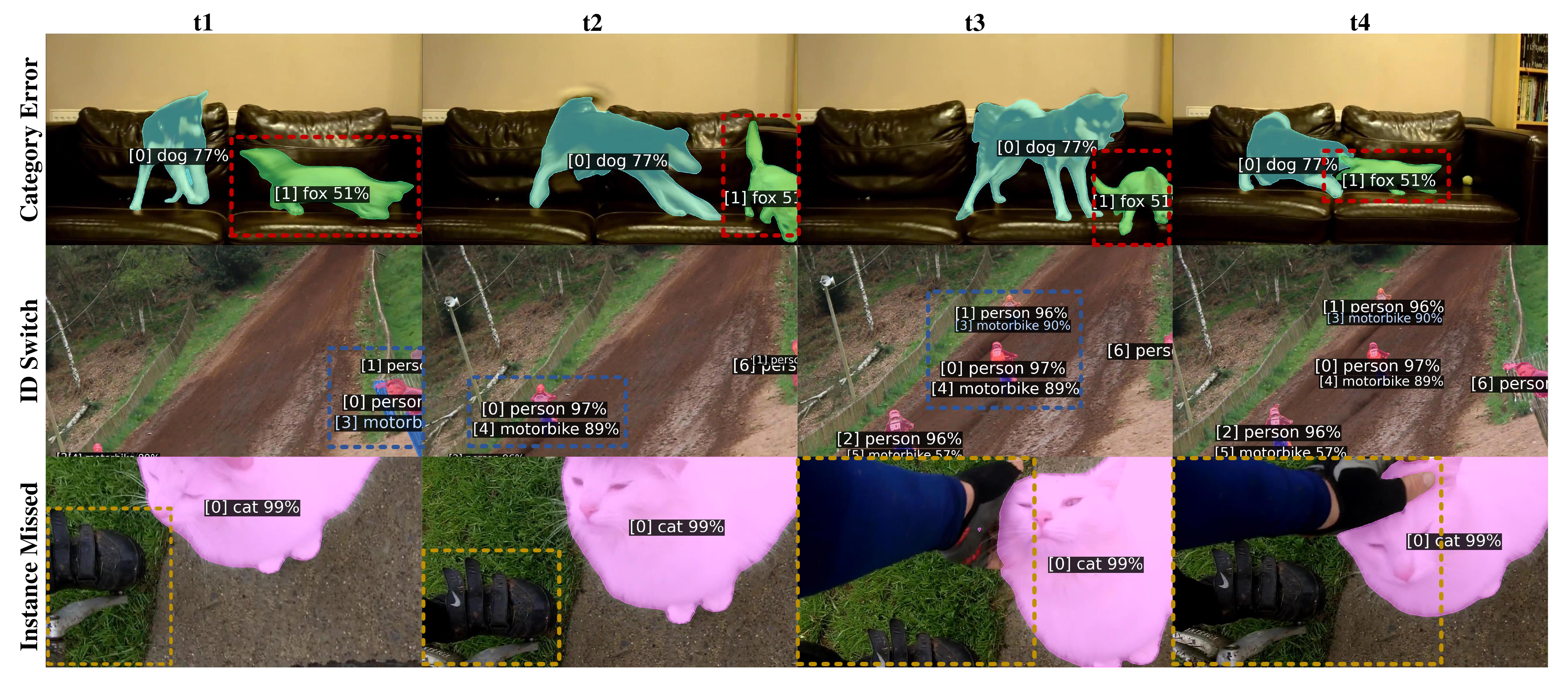}
    \caption{Failure cases on YouTubeVIS-2019 \cite{vis}.}
    \label{fig:failure_cases}
    \vspace{-3mm}
 \end{figure}

%% file: tex/tab_instseg.tex
\begin{table}[!htbp]
  \centering
  \vspace{-2mm}
    \caption{Performance comparison between image vs. video pre-training. 
    }
    \label{tab:instseg}%
    \begin{tabular}{lccccc}
    \toprule
    Methods & Archi.   &  VP& AP          & AP$_{\mathtt{50}}$        & AP$_{\mathtt{75}}$       \\
    \midrule
     M2F \cite{mask2former} & &  & 43.7 & 66.0 & 46.9  \\
     M2F && \cellcolor{gray!30}{\checkmark}& \cellcolor{gray!30}{45.3} & \cellcolor{gray!30}{68.3} & \cellcolor{gray!30}{48.9} \\
MD \cite{maskdino}  & &   &46.3 & 69.0 & 50.7  \\
MD &\multirow{-4}{*}{\rotatebox{90}{\scriptsize ResNet50}}&\cellcolor{gray!30}{\checkmark}  & \cellcolor{gray!30}{47.2} & \cellcolor{gray!30}{70.2} & \cellcolor{gray!30}{51.8}   \\
    \midrule
    
 M2F \cite{mask2former} &  &   & 50.1 & - & - \\
M2F& \multirow{-2}[1]{*}{\rotatebox{90}{\scriptsize Swin-L}}&\cellcolor{gray!30}{\checkmark} & \cellcolor{gray!30}{50.5} & \cellcolor{gray!30}{74.9} & \cellcolor{gray!30}{54.9} \\
    \bottomrule
    \end{tabular}%
    \vspace{-5mm}
\end{table}%

%% file: tex/tab_pool.tex
\begin{table}[h]
  \caption{The augmentation pool used in CoSt.}
  \centering
\begin{tabular}{ll}
\toprule
    Identity &  Returns the original image.\\
    Autocontrast & Maximizes (normalize) the image contrast.\\
    Equalize & Equalize the image histogram.\\
    Gaussian blur & Blurs the image with a Gaussian kernel.\\ 
    Contrast  & Adjusts the contrast of the image by [0.05, 0.95].\\
    Sharpness  &  Adjusts the sharpness of the image by [0.05, 0.95].\\
    Color & Enhances the color balance of the image by [0.05, 0.95].\\
    Brightness &Adjusts the brightness of the image by [0.05, 0.95].\\
    Hue & Jitters the hue of the image by [0.0, 0.5].\\
    Posterize  &Reduces each pixel to [4, 8] bits.\\
    Solarize &Inverts image pixels above a threshold from [1, 256).\\
    Rotation & Roates the image by a angle from [-15, 15].\\
    \bottomrule
\end{tabular} \label{tab:pool}
\end{table}

%% file: tex/tab_insbaseline.tex
\begin{table}[h]
  \caption{Ablation experiments of performance were conducted using different video pre-trained models.}
  \label{tab:ins_baseline}
  \centering
  \vspace{-1mm}
    \begin{tabular}{lcc}
    \toprule
   Model & AP$^{\mathtt{COCO}}$ &AP$^{\mathtt{YV19}}$ \\
    \midrule
   Mask DINO~\cite{maskdino}  & \cellcolor{gray!30}\textbf{47.2}&60.5    \\
     Mask2Former~\cite{mask2former} & 45.3& \cellcolor{gray!30}\textbf{61.8} \\
    \bottomrule
      \end{tabular}
\end{table}

%% file: tex/tab_coco.tex
\begin{table}
  \caption{Performance comparison on the COCO dataset with different temporal modules in the video pre-training phase.}
  \label{tab:mstt_pretrain}
  \centering
    \begin{tabular}{cccccc}
    \toprule
    Long-term & Short-term & AP & AP$^{\mathtt{s}}$ &AP$^{\mathtt{m}}$ &AP$^{\mathtt{l}}$    \\
    \midrule
    - & - & 43.7 & 23.4  & 47.2 & 64.8                     \\
    $\checkmark$  & - & 44.6 & 24.3  & 48.5 & 65.5                     \\
    - & $\checkmark$  & 44.5 & 24.4  & 48.3 & 64.8                     \\
    \rowcolor{gray!30}  
    $\checkmark$&  $\checkmark$   &    \textbf{44.7} & \textbf{24.7}  & \textbf{48.3} &\textbf{64.9}  \\
    \bottomrule
    \end{tabular}%
\end{table}